\documentclass{article}

\usepackage[a4paper, total={6in, 9in}]{geometry}
\usepackage[utf8]{inputenc} % allow utf-8 input
\usepackage[T1]{fontenc}    % use 8-bit T1 fonts
\usepackage{hyperref}       % hyperlinks
\usepackage{url}            % simple URL typesetting
\usepackage{booktabs}       % professional-quality tables
\usepackage{amsfonts}       % blackboard math symbols
\usepackage{nicefrac}       % compact symbols for 1/2, etc.
\usepackage{microtype}      % microtypography
\usepackage{lipsum}
\usepackage{graphicx}
\usepackage{cite}
\usepackage{amsmath}
\usepackage{svg}
\usepackage{caption}
\usepackage{subcaption}
\usepackage{array}

\graphicspath{ {./images/} }
\title{A Robust, Open-Source Framework for Spiking Neural Networks on Low-End FPGAs}

\author{Andrew Fan, Simon D. Levy}
\date{}

\begin{document}

\maketitle
\begin{abstract}
As the demand for compute power in traditional neural networks has increased significantly, spiking neural networks (SNNs) have emerged as a potential solution to increasingly power-hungry neural networks. By operating on 0/1 spikes emitted by neurons instead of arithmetic multiply-and-accumulate operations, SNNs propagate information temporally and spatially, allowing for more efficient compute power. To this end, many architectures for accelerating and simulating SNNs have been developed, including Loihi, TrueNorth, and SpiNNaker. However, these chips are largely inaccessible to the wider community. Field programmable gate arrays (FPGAs) have been explored to serve as a middle ground between neuromorphic and non-neuromorphic hardware, but many proposed architectures require expensive high-end FPGAs or target a single SNN topology. This paper presents a framework consisting of a robust SNN acceleration architecture and a Pytorch-based SNN model compiler. Targeting any-to-any and/or fully connected SNNs, the FPGA architecture features a synaptic array that tiles across the SNN to propagate spikes. The architecture targets low-end FPGAs and requires very little (6358 LUT, 40.5 BRAM) resources. The framework, tested on a low-end Xilinx Artix-7 FPGA at 100 MHz, achieves competitive speed in recognizing MNIST digits (0.52 ms/img). Further experiments also show accurate simulation of hand coded any-to-any spiking neural networks on toy problems. All code and setup instructions are available at \href{https://github.com/im-afan/snn-fpga}{\texttt{https://github.com/im-afan/snn-fpga}}. 
\end{abstract}

\section{Introduction}

Large artificial neural networks (ANNs) are the backbone of modern artificial intelligence. However, with the growing computational demand due to these neural networks from  larger architectures, the training and deployment of AI models is becoming increasingly power-hungry \cite{patterson2021carbonemissionslargeneural}.  Furthermore, as Moore's law approaches its limit, the growing need for compute for modern AI algorithms is also reaching its limit. To this end, spiking neural networks (SNNs) have been investigated as a power-efficient and fast alternative to ANNs. Instead of using floating-point or integer neuron activations, spiking neural networks operate on 0/1 spikes over time. As such, they propagate information both temporally and spatially, as opposed to ANNs, which only propagate information spatially. The advantage of this is that using custom hardware, expensive multiply-and-accumulate (MAC) operations are essentially eliminated, allowing for much lower power consumption.

Multiple architectures have been proposed to accelerate SNN operations. For example, Intel's Loihi features a many-core processor with neuromorphic elements such as synapses and neurons. It was designed and verified in a custom chip to allow for low-power spiking neural network operations \cite{loihi}. Similarly, IBM's TrueNorth features a synaptic crossbar with neurons and synapses in crossbar intersections, allowing for a massively parallel and low-power computation of neural networks \cite{truenorth}. However, these solutions are all based on ASICs, and research on the deployment of SNNs is hard due to the cost and scarcity of these chips. 

To this end, research on field-programmable gate arrays (FPGAs) has been conducted to investigate how the massive parallelism of FPGAs can be applied to facilitate SNN operations. However, many papers focus on implementing SNNs on large FPGA areas with more than 100K logic cells available. Furthermore, papers that do focus on deployments of SNNs on small FPGAs often optimize their architecture on a single neural network topology, overlooking any-to-any architectures that may allow for more robust behavior.

\section{Background}

\subsection{The Leaky-Integrate and Fire Neuron}

Spiking neural networks (SNNs) are a type of neural network that is fundamentally different from artificial neural networks. Instead of using floating-point or integer activations, SNNs operate on discrete 0/1 spikes over time. They support the same connectivity as a traditional neural networks, but instead of using a traditional activation function such as ReLU, sigmoid, or tanh, SNNs use a brain-inspired neuron model. The most common neuron model is the leaky-integrate and fire (LIF) neuron model.

The LIF model is characterized by receiving spikes from post-synaptic neurons and accumulating them as a membrane potential. When the membrane potential reaches a certain threshold, the neuron fires, resetting this membrane potential. The neuron dynamics, including leaking and fire can be characterized by a first-order diferential equation and a firing mechanism:
$$\tau \frac{dU}{dt}=I-U \text{ if } U<V_{\theta}$$
$$\text{fire spike, } U=0 \text{ if } U\geq V_{\theta}.$$
Where $I$ is the total presynaptic input to the neuron, $U$ is the membrane potential, $V_{\theta}$ is the threshold for firing a spike, and $\tau$ is a time constant characterizing how quickly the neuron leaks. While $\tau$ and $V_{\theta}$ can be configured as trainable parameters in an SNN, they are commonly set to constant values. For simplicity, however, SNNs are typically computed and simulated in discrete timesteps instead of in continuous time. Therefore, instead of solving the above differential equation to apply neuron dynamics, we instead approximate the leaking mechanism using Euler's method through time: $$U(t)=\beta\cdot U(t-1)+I(t) \text{, where } 0<\beta<1$$
$$y(t+1)=\Theta(U(t)-V_{\theta}).$$
In this case, $\Theta$ is the Heaviside step function, which outputs $0$ for negative inputs and $1$ otherwise. In practice, the LIF model can be treated as an activation function in a typical neural network.

\subsection{Backpropagation Through Time and Surrogate Gradient Descent}

To train an SNN using backpropagation, we treat it as a recurrent neural network, unrolling its operations out through time. At each timestep, we take the gradient of the loss of the output at that timestep with respect to the weights:
$$\frac{\partial U}{\partial W}=\sum_{t=0}^{T} \frac{\partial L}{\partial y(t)} \frac{\partial y(t)}{\partial U(t)} \frac{\partial U(t)}{\partial W}$$

The existence of the step function in the LIF model presents a fundamental difficulty in using traditional backpropagation methods with SNNs. Because the derivative of the Heaviside step function is $0$ everywhere except for when the input is $0$, the $\frac{\partial y(t)}{\partial U(t)}$ term is always $0$. The workaround is to smooth out the Heaviside step function during training, by replacing the step function with a surrogate function \cite{snntorch}. One common surrogate used is the $\text{arctan}$ surrogate:
$$\Theta(U)\approx \frac{1}{\pi}\arctan(\pi U)$$
$$\frac{\partial \Theta}{\partial U} \approx \frac{1}{\pi}\cdot\frac{1}{1+(\pi U)^2}.$$
By smoothing out the step function, gradients are allowed to flow through the weights, and the entire SNN can be successfully trained.

\subsection{Spiking Neural Networks on FPGAs}
Implementations of fast, low-power spiking neural networks on FPGAs have been explored because of the FPGA's potential for massive parallelism and ability to be reconfigured for different SNN architectures. A diverse range of architectures has been explored; for example, spiking convolutional neural networks (CNNs) have been an area of great interest due to their potential in accelerating neuromorphic image processing techniques. For example, Firefly \cite{firefly} greatly accelerated large spiking convolutional neural network computations by accelerating synaptic crossbars using digital signal processing blocks (DSP) on Xilinx FPGAs. This allowed for high-performance inference on convolutional neural networks. Firefly was designed for large-scale FPGA architectures; hundreds of DSP blocks and tens of thousands of LUTs were used in its implementation on Xilinx's Ultrascale FPGA. Similarly, SyncNN \cite{syncnn} explored multiple memory access and computation optimizations for FPGAs to design a highly optimized compute architecture for CNNs, allowing for state-of-the-art inference performance on multiple CNN architectures such as VGG and LeNet.

Resource-constrained implementations of fully connected SNNs on low-end FPGAs have also been explored. \cite{fcsnnfpga} trained a 784-1024-1024-10 spiking neural network on the MNIST dataset and designed an efficient architecture based on event queues for spike events in multiple time steps. The SNN weights and state were stored on DRAM.  This allowed for efficient inference (about 160 frames/sec) on the MNIST dataset with relatively low resource usage (5381 LUT, 7309 FF, 40.5 BRAM). Spiker was also designed for low-power, high-performance SNN computations \cite{spiker}. Spiker trained a 768-400 fully connected neural network using spike-time dependence plasticity (STDP) and took advantage of nonspiking action during neural network inference, allowing for ultra-efficient computations. Spiker+ \cite{spikerplus} improved this architecture by allowing backpropagation through time training using SNNTorch; it also included an open-source library for developers and researchers to train and deploy their own SNNs using the Spiker+ architecture. Other recent work in this area includes the TeNNLab RISP neuroprocessor \cite{risp} supporting any-to-any connections, and the NeuroCoreX processor developed at Oak Ridge National Laboratories \cite{10386808} supporting all-to-all connectivity.  The work we present below aims to contribute to these efforts by introducing a complete FPGA-based framework for both fully-connected and any-to-any SNN topologies.

\section{Architecture}
\subsection{Overview}
Off-chip, the neural network is trained and split into "tiles," which are chunks of the SNN with 16 pre-synaptic neurons and 16 post-synaptic neurons, for a maximum of 256 synapses. The tiles are ordered in memory so that they are sorted by the index of the 16 postsynaptic neurons, allowing for more efficient updates. With 8-bit quantization of neuron membrane potentials and weights, each tile has a size of 128 bitsfor post-synaptic membrane potentials and 2048 bits for weights. 

The SNN accelerator architecture consists of a synaptic crossbar, multiple LIF neuron units, and a control unit. The control unit synchronously loads the states of each tile from the FPGA block memory. For each tile, the synaptic crossbar then uses the loaded spikes and weights to compute the accumulated presynaptic potentials for the LIF units. The LIF units then accumulate the output spikes from the synaptic array and send spikes when their thresholds are reached. Outside of the architecture, a soft-core microprocessor is used to interface with a host computer by reading from the spike output BRAM and writing to the weight/input BRAM. Note that this computation method is strictly synchronous, meaning that there is no performance improvement from non-spiking. Overall, the processor has a throughput of 1 tile per clock cycle.

\begin{figure}[ht]
\centering
\captionsetup[subfigure]{justification=centering}
\begin{subfigure}[ht]{0.6\textwidth}
\centering
    \includegraphics[width=0.9\linewidth]{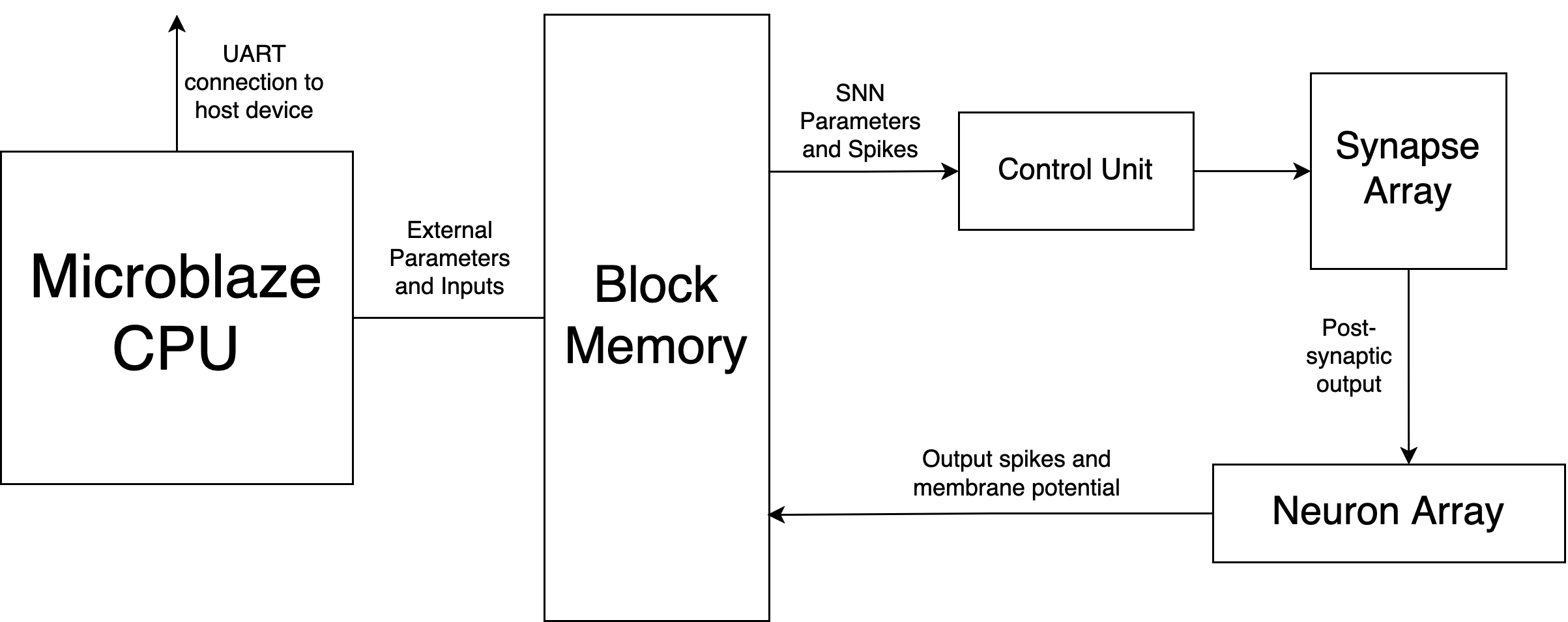}
    \centering
    \caption{High-level overview of SNN architecture, including neuron units, control unit, and synaptic crossbar.}
\end{subfigure}
\begin{subfigure}[ht]{0.35\textwidth}
\centering
\includegraphics[width=0.9\linewidth]{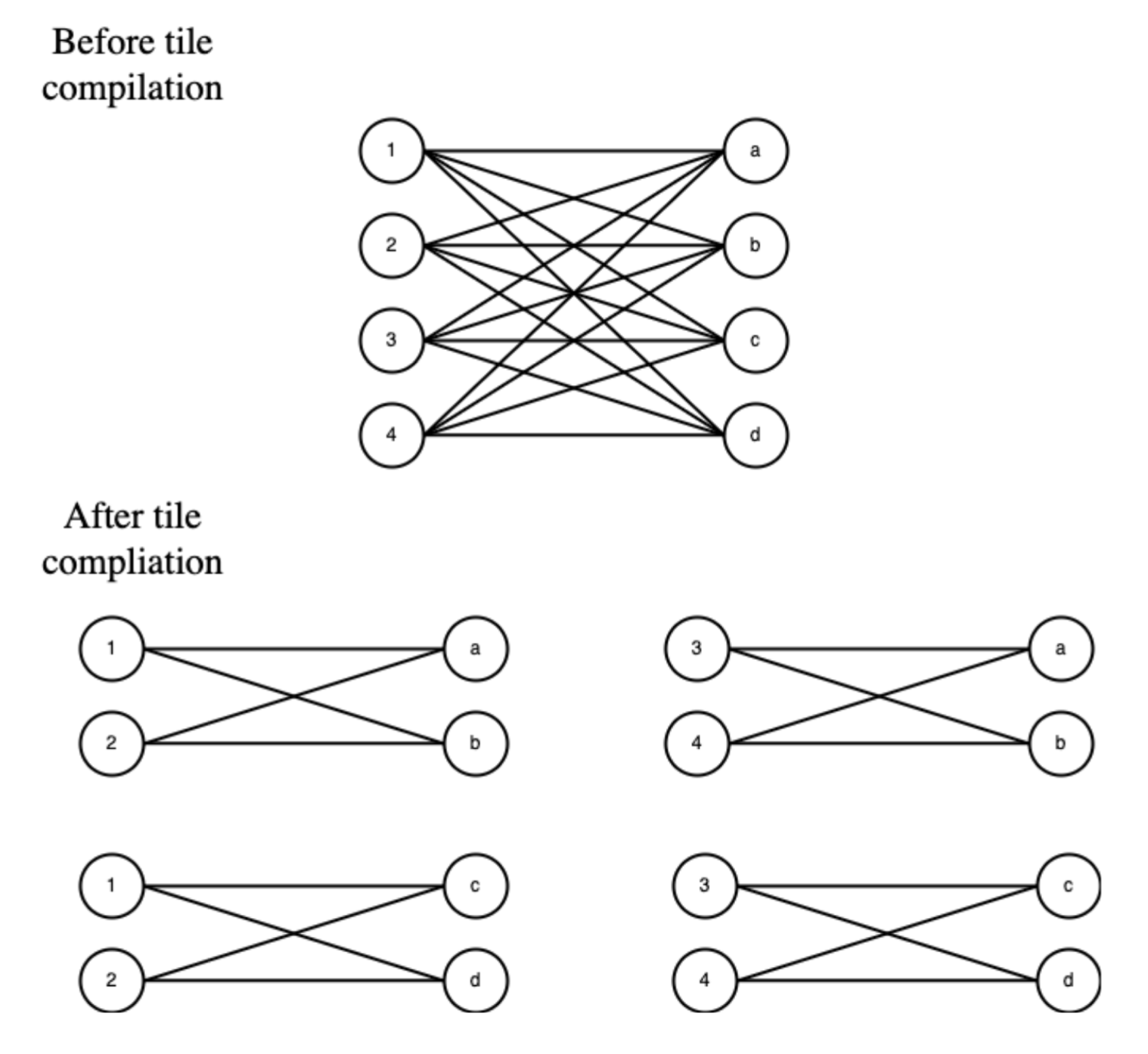}
\caption{Example of tiling using 2x2 tiles in a 4x4 fully-connected neural network.}
\label{fig:subim2}
\end{subfigure}
\caption{High-level components of the architecture.}
\end{figure}

A control unit streams the weights, neural network inputs, membrane potentials, and tile indices for each tile from block memory to the crossbar, with one tile being loaded per clock cycle. Since the address of the membrane potentials and neural network inputs depends on the tile indices, the tile indices must be loaded before some of the other data. Therefore, FIFOs are used to synchronize the data entering the synaptic crossbar. A RESET signal is pushed into another FIFO that propagates information through the synaptic array for the neuron array to prepare for a different output tile index (TILE\_IDX\_Y) when the value of TILE\_IDX\_Y is different from the previous value.

The synaptic crossbar contains a $16\times16$ weight matrix and receives 16 binary inputs, representing neuronal spikes. Using the weight matrix, it accumulates spikes and outputs 16 integers, representing the accumulated sums that should be added to each membrane potential. Furthermore, each column of the synaptic crossbar is a pipelined adder tree, which means that the synaptic crossbar has a throughput of one tile per clock cycle. Finally, the membrane potential increases, tile indices, and RESET signals are pushed into a FIFO, which is read by the neuron array control unit.

The neuron array takes in the accumulated spikes and increases the stored membrane potential accordingly.  When a RESET signal is received, the stored membrane potential is compared to the threshold; if the membrane potential is above the threshold, the neuron outputs a spike, and the output membrane potential is zero. Otherwise, the neuron does not emit a spike, and the output membrane potential is saved. The state of the output neurons in the tile is written to the respective BRAM unit; the spiking outputs are stored as a 16-bit binary number, and the membrane outputs are stored as 16 eight-bit signed integers. Finally, the stored membrane potential register is reset to zero. Note that there is no leak mechanism; this is to uphold hardware simplicity and to ensure minimum differences between training on decaying floating-point numbers and model deployment on integers. 

Furthermore, if the output neuron tile is designated as part of an output layer, it is written to a special output BRAM, addressed using both the simulation timestep and its index. In this architecture, the simulation timestep is at most 128 and the number of neuron output layer tiles is limited to a small number.

Input encoding is done with direct injection of voltage into the membrane potential of input neurons; that is, the input itself is not explicitly encoded by hardware. Instead, the input is directly added to the input neuron's membrane potential; the neuron spikes whenever the accumulated input values are high enough. When there is a large number of simulation timesteps, this is equivalent to a rate code.

\begin{figure}[ht]
\centering
\captionsetup[subfigure]{justification=raggedright}
\begin{subfigure}[ht]{0.75\textwidth}
\centering
\includegraphics[width=0.9\linewidth]{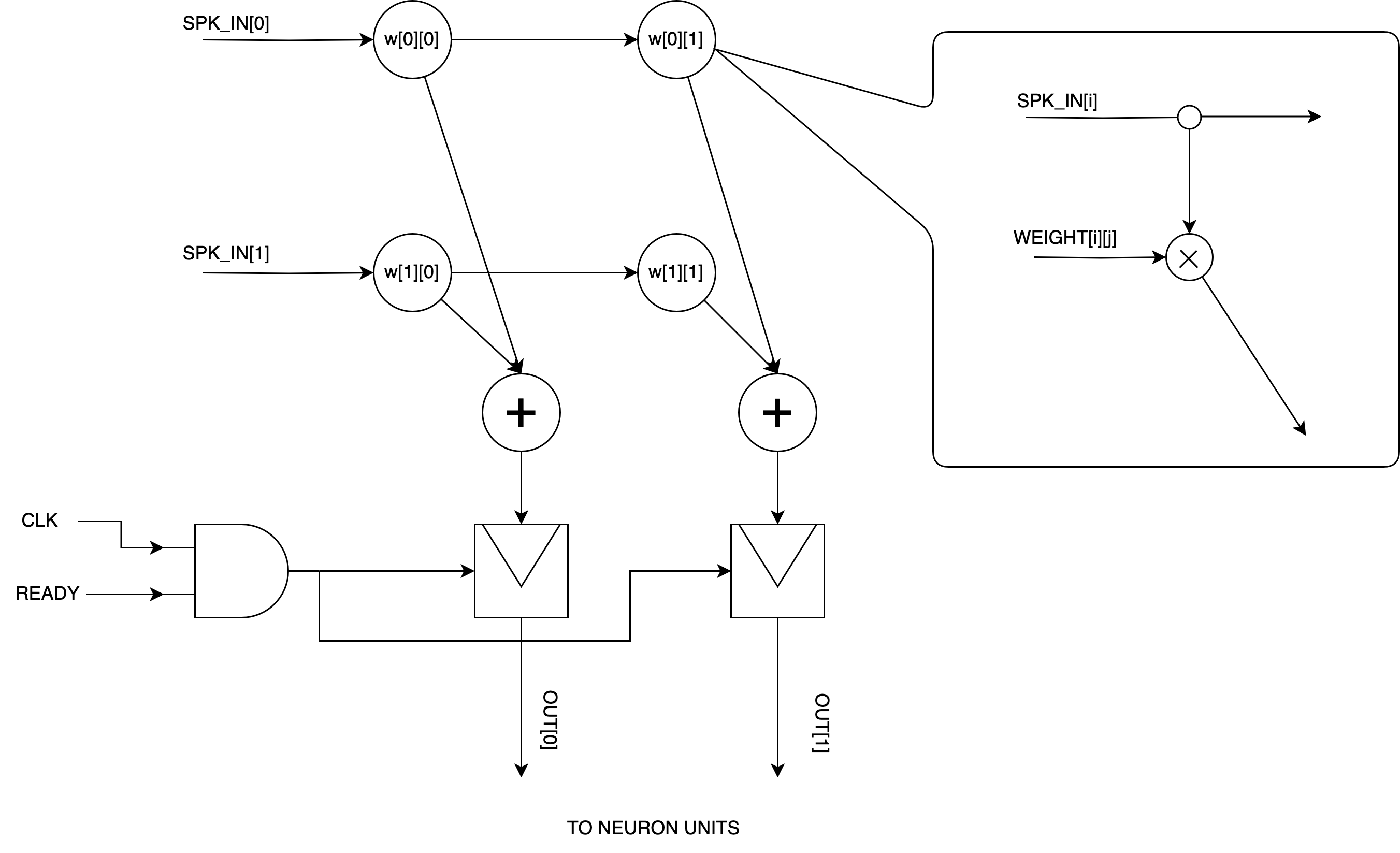} 
\caption{A 2x2 synaptic array. The synapses (units in the array) take in a weight and a 0/1 spike activation. Then, the synaptic output is fed into an adder tree with a delay of 1 clock and sent to the accumulated spikes.}
%\label{fig:subim1}
\end{subfigure}
\\
\begin{subfigure}[ht]{0.75\textwidth}
\centering
\includegraphics[width=0.9\linewidth]{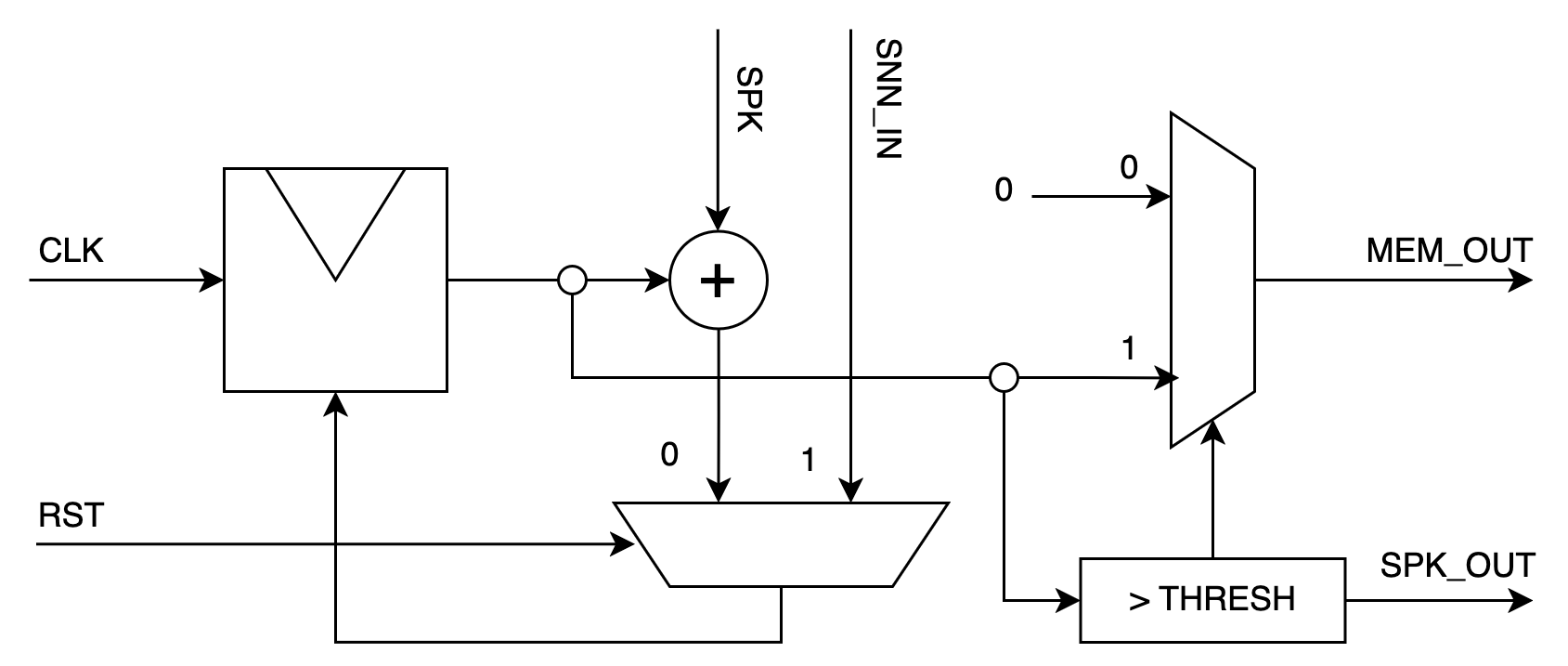}
\caption{Integrate and fire neuron used in FPGA architecture. For simplicity both training and hardware, there is no leak factor.}
\end{subfigure}
\caption{Synaptic array circuit (top) and IF neuron circuit (bottom).}
\end{figure}

\section{Experiments}
\subsection{Software Framework}
The Github repository \texttt{https://github.com/im-afan/snn-fpga} offers a library to compile, train, and quantize neural networks compatible with the architecture. Neural networks are trained on datasets using SNNTorch with no-leak IF neurons with a synaptic delay of one timestep. The SNNTorch training uses backpropagation through time with surrogate gradient descent. Currently, only fully connected neural networks are supported; any-to-any neural networks must be manually generated. The model compilation takes in the trained floating-point neural network, quantizes all weights to eight bits (which is sufficient for accuracy due to the threshold behavior of SNNs), then compiles the neural network into tiles using a method similar to Figure 1b. Finally, the tiles are loaded into either a C array deployable as high-level Microblaze code or as a .mem memory initialization file for synthesis.

\subsection{MNIST}
To test this architecture on large neural networks, a fully-connected neural network was trained off-chip on the MNIST dataset. The MNIST dataset is a set of 60,000 28x28 grayscale handwritten digits \cite{mnist}. The neural network was then deployed on the chip by modifying the BRAM initialization files. A Microblaze CPU was also set up on the FPGA chip to communicate via UART with the host computer to show the SNN output spikes every timestep. All testing was done on a low-end Digilent Basys3 FPGA board, which has limited logic cell resources. All hardware design was written in SystemVerilog and synthesized and deployed using Vivado. One challenge posed by using the Basys3 is the lack of external RAM; therefore, the accelerator only uses block memory to hold neural network states and weights, greatly limiting the maximum size of supported SNNs. The inference time and energy was determined using Verilog simulation in Vivado.
\begin{figure}[ht]
\centering
\begin{tabular} {| m{0.25\linewidth} | m{0.1\linewidth} |}
    \hline
    Clock (MHz) & 100  \\   \hline
    Weight Precision & int8 \\ \hline
    Avail. BRAM Blocks & 50 \\ \hline    
    Used BRAM Blocks (SNN) & 40.5 \\ \hline
    Used BRAM Blocks (CPU) & 4 \\ \hline
    Avail. LUT Cells & 20800 \\ \hline
    Used LUT Cells (SNN) & 6358  \\ \hline
    Used LUT Cells (CPU)  & 7106 \\ \hline
    NN Architecture & 784-128-10 \\   \hline
    Number of synapses & 101632\\ \hline
    T/img (ms) & 0.52\\ \hline    
    Power (W) & 0.381\\ \hline
    Energy/img (mJ) & 0.198\\ \hline
    Energy/syn (nJ) & 1.95\\ \hline
\end{tabular}
\caption{Results of testing this architecture on a 784-128-10 fully-connected spiking neural network.}
\end{figure}

\subsection{Toy Problems With Any-to-Any Connections}
To demonstrate the FPGA architecture's robustness towards non-traditional neural network architectures, a toy problem was used with a hand-coded any-to-any SNN: given a temporal input spike train with 2 spikes, determine whether the number of timesteps between the spikes is odd or even. The neuron connections and weights for the SNN were hand coded and compiled into tiles using the software framework, then tested on the FPGA. The results were accurate to the expected behavior of the hand-coded SNN.
\begin{figure}[ht]
\centering
\captionsetup[subfigure]{justification=centering}
\begin{subfigure}[ht]{0.45\textwidth}
\centering
\includegraphics[width=1\linewidth]{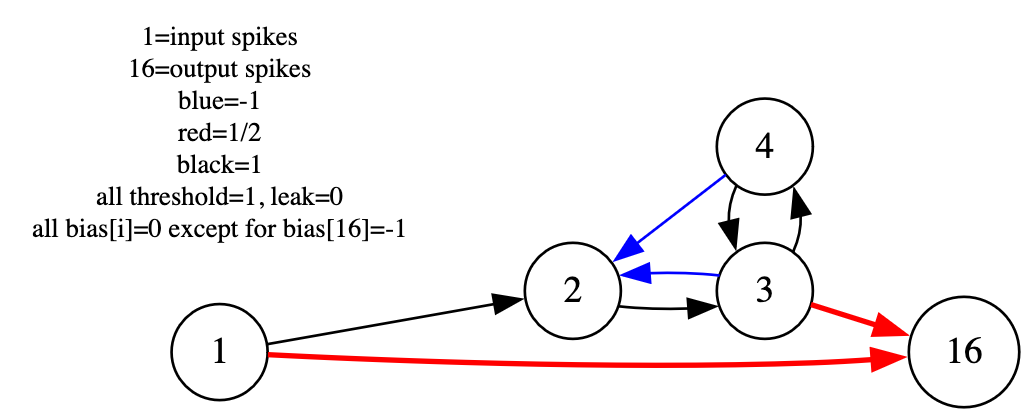} 
\caption{Recurrent, any-to-any SNN for solving the toy problem. The output neuron is indexed at 16 to demonstrate using multiple tiles for any-to-any SNNs. When the first input spike arrives, it passes through neuron 2, beginning an infinite parity counter in neurons 4 and 3, which prevents neuron 2 from spiking again. When the input spike arrives, it passes through an AND gate with neurons 1 and 3 to determine the parity of the time between the spikes.}
%\label{fig:subim1}
\end{subfigure}
\hspace{0.05\textwidth}
\begin{subfigure}[ht]{0.45\textwidth}
\centering
\begin{minipage}{\linewidth}
\begin{tabular}[ht]{| m{0.1\linewidth} | m{0.1\linewidth} | m{0.1\linewidth} | m{0.1\linewidth} |m{0.1\linewidth} | m{0.1\linewidth} |}
    \hline
    t\textbackslash idx & 1 & 2 & 3 & 4 & 16\\
    \hline
    t=1 & 1 & 0 & 0 & 0 & \textbf{0}\\
    \hline
    t=2 & 0 & 1 & 0 & 0 & \textbf{0}\\
    \hline
    t=3 & 1 & 0 & 1 & 0 & \textbf{0}\\
    \hline
    t=4 & 0 & 0 & 0 & 1 & \textbf{1}\\
    \hline
\end{tabular}
\end{minipage}
\vspace{0.2cm}

\begin{minipage}{\linewidth}
\begin{tabular}[ht]{| m{0.1\linewidth} | m{0.1\linewidth} | m{0.1\linewidth} | m{0.1\linewidth} |m{0.1\linewidth} | m{0.1\linewidth} |}
    \hline
    t\textbackslash idx & 1 & 2 & 3 & 4 & 16\\
    \hline
    t=1 & 1 & 0 & 0 & 0 & \textbf{0}\\
    \hline
    t=2 & 0 & 1 & 0 & 0 & \textbf{0}\\
    \hline
    t=3 & 0 & 0 & 1 & 0 & \textbf{0}\\
    \hline
    t=4 & 1 & 0 & 0 & 1 & \textbf{0}\\
    \hline
    t=5 & 0 & 0 & 1 & 0 & \textbf{0}\\
    \hline
\end{tabular}
\end{minipage}
\caption{Results of testing the any-to-any SNN architecture. When the input spikes (index 1) are 2 (even) timesteps apart, a spike is outputted. When the input spikes are 3 (odd) timesteps apart, no spikes are outputted.}
\end{subfigure}
\end{figure}

\subsection{Discussion}
This work proposes a robust and versatile FPGA implementation architecture for spiking neural network computations. Designed to be compatible with low-end FPGA architectures with few memory blocks, logic cells, and DDR RAM, the architecture proposed is lightweight (using only 13K LUT cells and 44.5 BRAM blocks when the soft-core CPU is used) and fast (requiring 520$\mu s$ to simulate a 784-128-10 SNN for 100 timesteps), making it comparable to similar architectures on fully-connected neural networks. The architecture includes a synaptic array, an array of simple IF neuron units, and a control unit that controls data flow to and from block memory. Beyond that, the architecture comes with a simple Python framework using SNNTorch to automatically train, quantize, and deploy models. 

Furthermore, the architecture has been proven to perform accurately on any-to-any spiking neural networks. This opens possibilities for deploying hand-coded logic gates and other neuromorphic algorithms on FPGAs. Likewise, it is also possible to deploy  any-to-any SNNs trained using bio-inspired or evolutionary algorithms. Furthermore, the ability to hand-code neural networks combined with the framework's ability for fully connected SNNs presents an interesting application. Using the tiling framework and FPGA architecture, future research may look to "chain" multiple neural networks together using both fully connected and simple hand-coded neural networks in series. This may allow fully neuromorphic systems in robotics and IoT to be deployed on a low-power FPGA.

Overall, this work creates a foundation for training and deploying simple spiking neural networks on small, cheap, and low-power FPGA architectures. This makes the framework very suitable for embedded FPGA applications in robotics, IoT, and more. Furthermore, this project is now an ongoing open source project available on Github, encouraging further research and use.

However, there are some limitations of the current framework. Since surrogate gradient descent is best applied for well-structured architectures like fully connected layers, usage on neural networks with backwards and any-to-any connections was limited to just hand-coded SNNs. Beyond any-to-any neural networks, future work may also investigate convolutional or other well-known architectures. Furthermore, hardware considerations also limited testing of the SNN processor architecture. Hardware testing was limited to the Basys3 FPGA, which lacks dedicated DDR RAM and features a low-end FPGA chip with less than 100K logic cells. Future work can focus on investigating how to scale this architecture to be more efficient on larger chips and make use of dedicated RAM to support larger models or on embedded FPGAs such as iCE40 FPGAs. Finally, the SNN processor architecture does not support asynchronous operation on spikes. That is, non-spiking behavior does not speed up SNN inference at all, which proves to be a very large slowdown when scaled to architectures with many neurons. Future work might explore ways to take advantage of the absence of spikes.

%\section{Methods}
%\section{Results}
\bibliographystyle{unsrt}  

%\begin{thebibliography}{1}
\bibliography{references}
%\end{thebibliography}

\end{document}